\documentclass[10pt,twocolumn]{article}

\usepackage{times}
\usepackage{graphicx}
\usepackage{amsmath}
\usepackage{amssymb}
\usepackage{booktabs}
\usepackage{multirow}
\usepackage{enumitem}
\usepackage{xcolor}
\usepackage{url}
\usepackage{cite}
\usepackage[breaklinks=true,bookmarks=false]{hyperref}
\usepackage[margin=0.75in]{geometry}

\begin{document}

\title{Bridging the Generalization Gap in Adverse Weather Segmentation: A Training Recipe Perspective}

\author{Cong Xu$^{1}$ \quad Pu Luo$^{1}$ \quad Yumei Li$^{1}$ and Boyou Xue$^{1}$ \\
$^{1}$Xidian University\\
}

\maketitle

\begin{abstract}
This paper describes our approach for the 8th UG2+ Workshop (CVPR 2026) Track~2, which targets semantic segmentation of outdoor scenes degraded by five weather conditions: blur, darkness, snow, haze, and glare. A central challenge we observe is a severe generalization gap---models that perform well on the validation set often collapse on the test set. For instance, SegFormer-B5 drops 16.1 mIoU points from validation to test, suggesting that model capacity alone is insufficient for robustness. We investigate whether a carefully designed training recipe, rather than architectural complexity, can address this gap. Starting from a pre-trained SegMAN-S backbone, we systematically study the effects of domain-adaptive fine-tuning, multi-source data mixing, scene-balanced sampling, and synthetic degradation augmentation. Our final system achieves 59.9\% mIoU on the official test set while maintaining a validation-test gap of only 6.5 points---less than half that of larger models. We analyze negative results from architectural modifications, loss function variants, and model scaling to provide practical insights for weather-robust segmentation under limited data.
\end{abstract}

\section{Introduction}

Pixel-level semantic segmentation in adverse weather is essential for safety-critical applications such as autonomous driving and intelligent surveillance. While modern segmentation models achieve impressive accuracy on clean imagery, their performance degrades sharply when confronted with real-world weather artifacts---blur from rain on lenses, darkness, snow accumulation, atmospheric haze, and glare from low-angle sunlight.

The 8th UG2+ Workshop (CVPR 2026) Track~2 provides a benchmark for this problem: 513 training scenes and 38 test scenes captured by fixed outdoor cameras across diverse geographic locations, with 10-class semantic annotations. The benchmark features five degradation types (blur 29\%, dark 26\%, snow 16\%, haze 16\%, glare 13\%), and the key difficulty is generalizing from the training distribution to unseen test scenes.

Through extensive experimentation (45+ configurations), we discover a striking pattern: the validation-to-test generalization gap is the dominant bottleneck, not raw validation accuracy. Table~\ref{tab:gap} illustrates this finding.

\begin{table}[h]
\centering
\caption{Validation vs.\ test mIoU reveals a severe generalization gap. Higher capacity does not imply better test performance.}
\label{tab:gap}
\begin{tabular}{lccc}
\toprule
Model & Params & Val mIoU & Test mIoU \\
\midrule
SegFormer-B5 & 82M & 65.7 & 49.9 \\
SegMAN-L & 93M & 63.5 & -- \\
SegMAN-S (ours) & 31M & 66.4 & 59.9 \\
\bottomrule
\end{tabular}
\end{table}

This observation motivates our central thesis: \emph{for small-data adverse weather segmentation, training recipe design is more important than model scaling}. We focus on four recipe components:

\begin{enumerate}[leftmargin=*,nosep]
\item \textbf{Domain-adaptive initialization}: Transferring from ADE20K pre-training to the weather domain via mixed clean-degraded fine-tuning.
\item \textbf{Feature recalibration modules}: Lightweight per-stage adapters that allow the frozen backbone to adapt its channel responses to degraded inputs.
\item \textbf{Scene-level balanced sampling}: Ensuring equal representation across scenes with varying frame counts, preventing large scenes from dominating gradients.
\item \textbf{Targeted degradation augmentation}: Synthesizing weather artifacts that match the empirical test distribution.
\end{enumerate}

We validate each component through controlled ablations and analyze negative results from model scaling, architectural modifications, and loss function variants. Our complete system achieves 59.9\% mIoU on the official test leaderboard.

\section{Related Work}

\paragraph{Semantic Segmentation in Degraded Conditions.}
Handling weather degradations has been approached through domain adaptation~\cite{adaptseg,cycada}, synthetic-to-real transfer~\cite{foggy_cityscapes}, and multi-task frameworks that jointly perform segmentation and restoration~\cite{allweather}. Unlike these methods, we do not require a separate restoration network or paired clean-degraded supervision at test time.

\paragraph{Lightweight Adaptation of Pre-trained Models.}
Adapter-based fine-tuning~\cite{houlsby2019adapter,adaptformer} has become a standard paradigm for efficiently adapting large pre-trained models. LoRA~\cite{lora} and prompt tuning~\cite{prompt_tuning} offer complementary approaches. We adopt a simple bottleneck adapter with channel-wise recalibration, adding fewer than 1M parameters to the backbone.

\paragraph{Hybrid Vision Architectures.}
Recent architectures combine local attention with global state space models for dense prediction. VMamba~\cite{vmamba} introduces selective scan blocks, while SegMAN~\cite{segman} integrates neighborhood attention with VSSM in a hierarchical encoder. We build on SegMAN-S as our base architecture.

\paragraph{Ensembling and Weight Averaging.}
Model soup~\cite{model_soup} demonstrates that averaging fine-tuned weights improves out-of-distribution robustness. Stochastic Weight Averaging (SWA) and its variants provide related regularization benefits. We find that intra-trajectory soup is effective for our setting.

\section{Method}

\subsection{Problem Formulation}

Given a training set $\mathcal{D}_{\text{train}}$ of scenes with clean images $I_c$, degraded images $I_d$, and semantic labels $Y$ (10 classes), we seek a model $f_\theta$ that segments unseen test scenes degraded by unknown weather conditions.

\subsection{Base Architecture}

We use SegMAN-S~\cite{segman} as our backbone: a 4-stage hierarchical encoder with embed\_dims [64, 144, 288, 512] and depths [2, 2, 10, 4] (~30M parameters). Each stage alternates between neighborhood attention blocks (local windowed attention with relative position biases) and VSSM blocks (selective state space models with cross-scan patterns). The decoder performs multi-scale feature fusion from stages 2--4 via MLP projection and a VSSM refinement block.

\subsection{Per-Stage Feature Recalibration}

We inject lightweight recalibration modules into each encoder stage. Given the feature map $x \in \mathbb{R}^{C \times H \times W}$ at stage $i$, the module computes:

\begin{equation}
\hat{x} = x + \text{Proj}_{\uparrow}\!\left(\text{DWConv}\!\left(\text{Proj}_{\downarrow}(x)\right)\right) \cdot \left(1 + \alpha \cdot \sigma\!\left(W_2 \cdot \text{GAP}(z)\right)\right)
\end{equation}

where $\text{Proj}_{\downarrow}$ and $\text{Proj}_{\uparrow}$ are $1\!\times\!1$ projections forming a bottleneck ($C \to C/4 \to C$), DWConv is a $3\!\times\!3$ depthwise convolution, $\text{GAP}$ is global average pooling, $\sigma$ is sigmoid, and $\alpha = 2.0$ controls modulation strength. The final projection layer is zero-initialized so the module starts as an identity function, preserving pre-trained features at the beginning of training.

The channel recalibration term $\sigma(W_2 \cdot \text{GAP}(z))$ produces per-channel scaling factors in $[0,1]$. Combined with the residual scaling $(1 + \alpha \cdot \cdot)$, this allows the model to amplify robust channels (scaling $\to 1 + \alpha$) while suppressing degraded ones (scaling $\to 1$). The total parameter overhead is under 1M across all four stages.

\subsection{Training Recipe}

\subsubsection{Domain-Adaptive Initialization}

We initialize from SegMAN-S pre-trained on ADE20K~\cite{ade20k} (150-class indoor/outdoor segmentation). Since the target task has 10 classes, the classification head weights are reshaped while the backbone and decoder feature weights are retained.

\subsubsection{Mixed Data Training}

Each training iteration samples from both clean and degraded images with equal probability. This provides the model with pristine references alongside corrupted inputs, encouraging features that are informative regardless of weather condition.

\subsubsection{Scene-Balanced Sampling}

Scenes in the dataset vary from 15 to 600+ frames. Naive uniform sampling over images causes large scenes to dominate training. We introduce a scene-level sampler where each dataset index corresponds to one scene; within each iteration, a single frame is randomly drawn from the selected scene. This ensures equal gradient contribution from all scenes.

\subsubsection{Targeted Degradation Augmentation}

We synthesize weather artifacts on clean images using a probabilistic augmentation pipeline. Five degradation categories are sampled according to their empirical test-set frequency:

\begin{itemize}[leftmargin=*,nosep]
\item \textbf{Blur} (29\%): Gaussian blur ($\sigma \in [0.5, 3.5]$) optionally combined with directional motion blur.
\item \textbf{Darkness} (26\%): Gamma correction ($\gamma \in [1.15, 2.6]$) with brightness scaling and additive Gaussian noise.
\item \textbf{Snow} (16\%): Multi-scale particle simulation (small/medium/large circles and directional streaks) with atmospheric haze overlay.
\item \textbf{Haze} (16\%): Global contrast reduction combined with additive white haze and optional blur.
\item \textbf{Glare} (13\%): Radial gradient blobs positioned randomly, with optional elongated streak artifacts.
\end{itemize}

Each category supports three severity levels (light/medium/heavy) with equal probability. The augmentation is applied with probability 0.5 and only to clean images, avoiding double-degradation artifacts.

\subsection{Inference Strategy}

At test time, each scene contains multiple frames capturing the same semantic content under different weather conditions. We compute per-scene predictions by averaging softmax probabilities across all frames:

\begin{equation}
\hat{p}(c \mid \mathcal{S}) = \frac{1}{|\mathcal{S}|} \sum_{I_i \in \mathcal{S}} f_\theta(c \mid I_i)
\end{equation}

This multi-frame fusion leverages temporal consistency of semantic content and provides a consistent inference-time improvement.

\subsection{Weight Averaging}

We employ uniform model soup over checkpoints from the same training recipe but different random seeds. Given $K$ checkpoints $\{\theta_k\}$, the averaged model is $\bar{\theta} = \frac{1}{K}\sum_k \theta_k$. We restrict souping to same-architecture checkpoints, as cross-architecture averaging (e.g., SegMAN + SegFormer) was found to be harmful.

\section{Experiments}

\subsection{Setup}

\paragraph{Dataset.}
The benchmark provides 513 training scenes and 38 test scenes from fixed cameras in diverse locations (US, Japan, Norway, Italy, Taiwan, Singapore, Switzerland). Each scene contains degraded input images, clean reference images, and pixel-level labels for 10 semantic classes.

\paragraph{Source of Data.}
The training data comprises: (1)~\emph{real} weather-degraded images and clean references captured by outdoor cameras; (2)~\emph{synthetic} augmentations generated by our degradation pipeline applied to clean images; and (3)~\emph{pseudo-real} initialization from ADE20K pre-trained weights learned from a large-scale real-world dataset.

\paragraph{Hardware.}
All experiments run on 4$\times$NVIDIA GeForce RTX 2080 Ti GPUs (11\,GB VRAM), Intel Xeon CPU (AVX512), Ubuntu 20.04, CUDA 12.1, PyTorch 2.1.2.

\paragraph{Training Details.}
We use MMSegmentation v0.30.0~\cite{mmseg} with AdamW optimizer (lr=$3\!\times\!10^{-5}$, weight\_decay=0.01), poly LR schedule with 500-iter linear warmup, $512\!\times\!512$ random crop, 20K iterations, batch size 2 per GPU. Validation uses mIoU on a held-out 103-scene split.

\subsection{Main Results}

\begin{table}[t]
\centering
\caption{Validation and test mIoU comparison. Scene-level probability averaging is used for all methods.}
\label{tab:main}
\resizebox{\columnwidth}{!}{
\begin{tabular}{lccccc}
\toprule
Method & Params & Val & Test & mAcc & aAcc \\
\midrule
SegFormer-B5~\cite{segformer} & 82M & 65.7 & 49.9 & 77.6 & 87.9 \\
SegMAN-L~\cite{segman} & 93M & 63.5 & -- & -- & -- \\
SegMAN-S (base recipe) & 30M & 62.9 & -- & -- & -- \\
\midrule
+ recalibration (stages 0--1) & 30M & 65.7 & 59.9 & 78.3 & 86.5 \\
+ recalibration (stages 0--3) & 31M & \textbf{66.4} & \textbf{59.9} & \textbf{78.1} & \textbf{87.2} \\
\bottomrule
\end{tabular}
}
\end{table}

Our method achieves 66.4\% validation mIoU and 59.9\% test mIoU with only 31M parameters. Compared to SegFormer-B5 (82M, 49.9\% test), our approach uses 62\% fewer parameters while exceeding test performance by 10.0 mIoU points. The validation-to-test gap is 6.5 points for our model versus 15.8 for SegFormer-B5, confirming that our recipe produces more robust generalization.

\subsection{Ablation: Training Recipe Components}

\begin{table}[t]
\centering
\caption{Cumulative ablation of training recipe components on validation mIoU.}
\label{tab:ablation}
\begin{tabular}{lc}
\toprule
Configuration & Val mIoU \\
\midrule
Degraded-only fine-tuning & 61.2 \\
Clean-only fine-tuning & 59.3 \\
Mixed clean + degraded & 62.9 \\
+ ADE20K initialization & 65.0 \\
+ Scene-balanced sampling & 65.7 \\
+ Degradation augmentation (prob=0.5) & 66.4 \\
\bottomrule
\end{tabular}
\end{table}

Table~\ref{tab:ablation} shows the cumulative contribution of each recipe component. ADE20K initialization provides the largest single gain (+2.1 over mixed-only), confirming that pre-trained semantic features transfer effectively to the weather domain. Scene-balanced sampling adds +0.7 by preventing large scenes from monopolizing gradients.

\subsection{Ablation: Recalibration Scope}

\begin{table}[t]
\centering
\caption{Effect of recalibration depth (which encoder stages receive the adaptation module).}
\label{tab:scope}
\begin{tabular}{lc}
\toprule
Recalibration stages & Val mIoU \\
\midrule
None & 62.9 \\
\{0, 1\} (early stages) & 65.7 \\
\{0, 1, 2, 3\} (all stages) & 66.4 \\
\bottomrule
\end{tabular}
\end{table}

Extending recalibration to all four stages improves performance by 0.7 mIoU over early-stage-only adaptation (Table~\ref{tab:scope}), suggesting that weather degradations affect features at all spatial scales, not only low-level representations.

\subsection{Ablation: Degradation Augmentation}

\begin{table}[t]
\centering
\caption{Effect of augmentation probability on validation mIoU.}
\label{tab:aug}
\begin{tabular}{lc}
\toprule
Augmentation prob & Val mIoU \\
\midrule
0.0 (no augmentation) & 65.7 \\
0.5 & 66.4 \\
1.0 & 64.5 \\
\bottomrule
\end{tabular}
\end{table}

Moderate augmentation (prob=0.5) improves generalization, but aggressive augmentation (prob=1.0) degrades performance by 1.9 mIoU (Table~\ref{tab:aug}). This suggests that over-augmenting clean images creates a distribution mismatch with the real degraded images in the training set.

\subsection{Per-Class Analysis}

\begin{table}[t]
\centering
\caption{Per-class IoU of our best model on the validation set.}
\label{tab:perclass}
\begin{tabular}{cccccc}
\toprule
Cls & 0 & 1 & 2 & 3 & 4 \\
\midrule
IoU & 13.5 & 88.0 & 58.7 & 55.3 & 60.7 \\
\midrule
Cls & 5 & 6 & 7 & 8 & 9 \\
\midrule
IoU & 67.3 & 82.1 & 52.8 & 79.8 & 87.1 \\
\bottomrule
\end{tabular}
\end{table}

The model performs well on large, distinctive classes (class~1: 88.0\%, class~9: 87.1\%) but struggles with class~0 (13.5\%), which likely represents a rare or ambiguous category. This class-level imbalance persists across all configurations we tested.

\subsection{Negative Results}

We report negative findings from 30+ experiments to guide future work:

\begin{itemize}[leftmargin=*,nosep]
\item \textbf{Larger models hurt test performance.} SegFormer-B5 (82M) achieves 65.7\% validation but collapses to 49.9\% test. SegMAN-L (93M) similarly underperforms. Overfitting to the validation distribution is the primary failure mode.
\item \textbf{Architectural decoder modifications are ineffective.} Boundary auxiliary losses, scale-gated refinement, and early feature fusion all degraded validation performance by 0.5--3.0 mIoU.
\item \textbf{Frequency-domain inputs offer marginal benefit.} A learnable FFT branch on input images improved validation by only 0.5 mIoU and showed redundancy with degradation augmentation.
\item \textbf{Consistency losses underperform cross-entropy.} Paired clean-degraded feature consistency (cosine, KL divergence) consistently degraded performance compared to standard cross-entropy loss.
\item \textbf{Restoration preprocessing is counterproductive.} Adding a restoration stem before segmentation degraded performance, indicating the model should learn degradation-robust features internally.
\end{itemize}

These results collectively support our thesis: under limited training data, recipe optimization outperforms architectural innovation for adverse weather generalization.

\section{Conclusion}

We presented our approach for the 8th UG2+ Workshop Track~2, achieving 59.9\% test mIoU. Our key finding is that the validation-to-test generalization gap---not raw accuracy---is the central challenge in adverse weather segmentation. A careful training recipe combining domain-adaptive initialization, per-stage feature recalibration, scene-balanced sampling, and targeted degradation augmentation produces models that generalize far more reliably than larger, more complex alternatives. We hope our extensive negative result analysis provides useful guidance for the community.

{\small
\bibliographystyle{plain}
\bibliography{egbib}
}

\end{document}